\documentclass[conference]{IEEEtran}
\IEEEoverridecommandlockouts
\usepackage{cite}
\usepackage{amsmath,amssymb,amsfonts}
\usepackage{algorithmic}
\usepackage{graphicx}
\usepackage{textcomp}
\usepackage{xcolor}
\usepackage{booktabs}
\usepackage{multirow}
\usepackage{url} 
\def\BibTeX{{\rm B\kern-.05em{\sc i\kern-.025em b}\kern-.08em
    T\kern-.1667em\lower.7ex\hbox{E}\kern-.125emX}}

\makeatletter
\newcommand{\linebreakand}{%
  \end{@IEEEauthorhalign}
  \hfill\mbox{}\par
  \mbox{}\hfill\begin{@IEEEauthorhalign}
}
\makeatother

\begin{document}

\title{Alternative Speech: Complementary Method to Counter-Narrative for Better Discourse\\
\thanks{This research was supported by Basic Science Research Program through the National Research Foundation of Korea(NRF) funded by the Ministry of Education(NRF-2021R1A6A1A03045425). This work was supported by Institute for Information \& communications Technology Planning \& Evaluation(IITP) grant funded by the Korea government(MSIT) (No. 2022-0-00369, (Part 4) Development of AI Technology to support Expert Decision-making that can Explain the Reasons/Grounds for Judgment Results based on Expert Knowledge). This research was supported by the MSIT(Ministry of Science and ICT), Korea, under the ITRC(Information Technology Research Center) support program(IITP-2023-2018-0-01405) supervised by the IITP(Institute for Information \& Communications Technology Planning \& Evaluation).}
}

\author{\IEEEauthorblockN{Seungyoon Lee$^{\dagger}$}
\IEEEauthorblockA{\textit{Computer Science and Engineering} \\
\textit{Korea University}\\
Seoul, Korea \\
dltmddbs100@korea.ac.kr}
\and
\IEEEauthorblockN{Dahyun Jung$^{\dagger}$}
\IEEEauthorblockA{\textit{Computer Science and Engineering} \\
\textit{Korea University}\\
Seoul, Korea \\
dhaabb55@korea.ac.kr}
\and
\IEEEauthorblockN{Chanjun Park$^{\dagger}$}
\IEEEauthorblockA{\textit{Large Language Model Team} \\
\textit{Upstage}\\
Gyeonggi-do, Korea \\
chanjun.park@upstage.ai}
\linebreakand 
\IEEEauthorblockN{Seolhwa Lee}
\IEEEauthorblockA{\textit{Computer Science} \\
\textit{Technical University of Darmstadt}\\
Darmstadt, Germany \\
whiteldark@gmail.com}
\and
\IEEEauthorblockN{Heuiseok Lim$^{\ast}$~\thanks{$^*$~Corresponding Authors, $^\dagger$~Equal Contributions}}
\IEEEauthorblockA{\textit{Computer Science and Engineering} \\
\textit{Korea University}\\
Seoul, Korea \\
limhseok@korea.ac.kr}
}

\maketitle

\begin{abstract}
\textit{Warning: This paper contains examples of stereotypes that may be offensive or upsetting.} \\[5pt]
We introduce the concept of ``Alternative Speech'' as a new way to directly combat hate speech and complement the limitations of counter-narrative. An alternative speech provides practical alternatives to hate speech in real-world scenarios by offering speech-level corrections to speakers while considering the surrounding context and promoting speakers to reform. Further, an alternative speech can combat hate speech alongside counter-narratives, offering a useful tool to address social issues such as racial discrimination and gender inequality. We propose the new concept and provide detailed guidelines for constructing the necessary dataset. Through discussion, we demonstrate that combining alternative speech and counter-narrative can be a more effective strategy for combating hate speech by complementing specificity and guiding capacity of counter-narrative. This paper presents another perspective for dealing with hate speech, offering viable remedies to complement the constraints of current approaches to mitigating harmful bias.
\end{abstract}

\begin{IEEEkeywords}
Hate Speech, Social Bias, Counter-Narrative
\end{IEEEkeywords}

\section{Introduction}
The conventional approach to mitigating the proliferation and influence of online hatred typically involves blocking or restricting users' expression, often relying on detection mechanisms~\cite{caselli2021hatebert,rottger-etal-2022-multilingual,baldini-etal-2022-fairness,alkhamissi-etal-2022-token}. Nonetheless, as it hinges on filtering through detection, this approach risks infringing on the right to freedom of expression and is not an ultimate remedy for diminishing the prevalence of hate speech~\cite{myers2018censored}. To tackle these limitations effectively, a counter-narrative is a viable solution to thwart the genesis of hate by enhancing users' prejudiced outlooks, furnishing factual information to challenge and rectify their discriminatory convictions, and discouraging the use of hateful language~\cite{milner2013counter,benesch2014countering,silverman2016impact}.

\begin{figure}[t]
\centerline{\includegraphics[width=.45\textwidth]{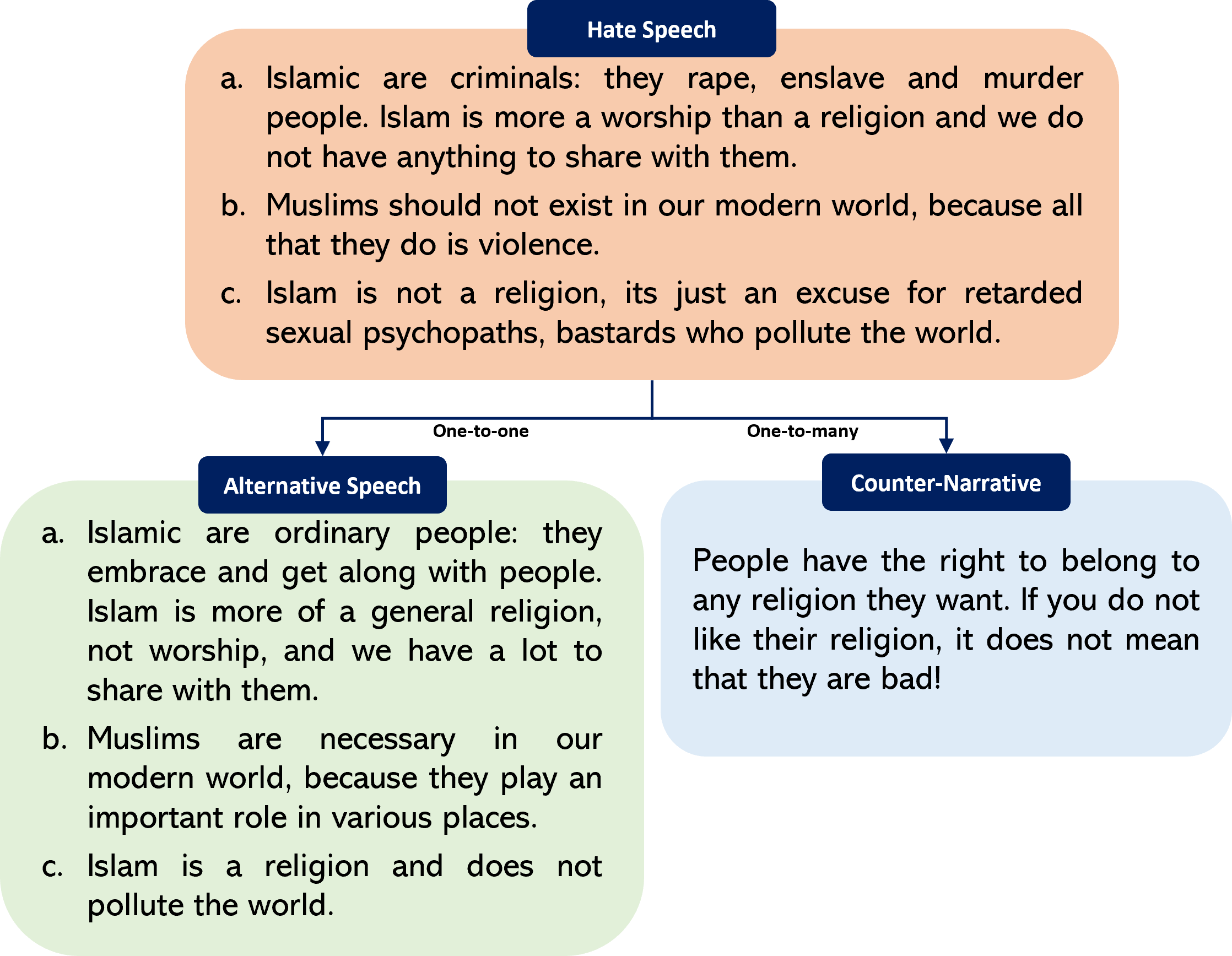}}
\caption{Examples of counter-narrative and alternative speech for hate speech. Given hate speech and counter-narrative are extracted from the CONAN dataset~\cite{chung2019conan}.}
\label{fig:exam_as}
\end{figure}

Counter-narrative is a fact-based argument that aims at suppressing hate speech by correcting discriminatory perceptions voiced with credible evidence~\cite{benesch2014countering,silverman2016impact}. While counter-narrative has shown promising results in reducing the impact of hate speech~\cite{schieb2016governing}, it suffers from two main limitations.

The first limitation of counter-narrative is their inability to furnish users with practical alternative expressions in real-world scenarios, as it aims to edify by pointing out with convincing evidence. Counter-narratives are typically structured around a central phrase in the hate speech or a response to the targeted group, which results in them featuring numerous one-to-many matching statements rather than one-to-one matches for a single utterance. In particular, while a counter-narrative can identify the inappropriateness of an utterance and lead to an appropriate rationale, it lacks direct instruction on the precise wording, which can be achieved by `substituting' the used utterance. In this context, a speech that focuses on rebuttal and lacks practical alternatives is unenlightening and may incur the risk of provoking users.

Another limitation of the counter-narrative is that it can mitigate the ethical concern about infringing on the speaker's freedom of expression compared to the detection method but cannot fully address it. Given the situation of countering hate speech in real-world scenarios, the counter-narrative returns a response visible only to that speaker. This approach may prevent the reproduction of hate speech, but in the end, the speech that occurred should be concealed through blocking or filtering. In terms of this practical implementation, it is difficult to circumvent the issue of over-blocking which silences the speaker by masking the entire sentence.

In this study, we propose ``Alternative Speech'', an utterance that directly purifies and corrects hatefulness, to overcome counter-narratives' abstruseness and radically mitigate rights restrictions in real-world scenarios. An alternative speech offers the speaker a practical alternative through sentence-level correction by providing a refined expression while retaining as much of the surrounding context as feasible. Furthermore, it goes beyond rebutting the speaker for the hateful language uttered and immediately replaces it with suitable wording, which is then displayed to online users, making it more applicable to real-world applications. It aims at enhancing the potential of enlightenment in a counter-narrative by directly correcting the speaker's aversion and compensating for the constraints of the counter-narrative.

We go beyond proposing alternative speech and provide detailed guidelines for building such data. Furthermore, we conduct a human survey to validate whether alternative speech can improve the speaker's biased perception. Our survey results demonstrate that alternative speech is more specific to the given context and target while functioning as a more efficacious approach to rectifying harmful statements. It is particularly effective when aligned with a counter-narrative since they play complementary roles and establish a feedback and correction loop.

\begin{table}[t]
\caption{Counter-narrative \& Alternative speech examples}
\begin{center}
\begin{tabular}{p{8.4cm}}
\toprule
\textbf{HS}: Muslims are asking too much! Their sharia has rules that cannot coexist with our culture.. They cannot ask us to not walk dogs in public! \\ \\
\textbf{CN}: You have to be careful about what you read! This is clearly a fake news, widespread to make a scandal. \\ \\
\textbf{AS}: Muslims are not asking for unreasonable demands! Their sharia has rules that can coexist with our culture.. No one cannot ask us not to walk dogs in public!
\\ \midrule 
\textbf{HS}: LGBTs just want to brainwash our children. They should not be taught about gender identity and sexual orientation in school. \\ \\
\textbf{CN}: Understanding gender diversity would break down stereotypes and reduce sexism. \\ \\
\textbf{AS}: LGBTs just want our children to learn about diversity. They should be taught about gender identity and sexual orientation in school.
\\ \midrule
\textbf{HS}: Their power to have movies banned and directors censured is proof that the Jews control Hollywood. \\ \\
\textbf{CN}: Censure and banning protect the vulnerable from seeing harmful content. Not such a bad thing, whoever does it. \\ \\
\textbf{AS}: Their power to have movies enriched and directors competent is proof that the Jews contribute to Hollywood.
\\ \midrule
\textbf{HS}: Blacks are a plague on society. They are vicious and cruel. They steal and are a menace to society. They corrupt the criminal justice system. They are basically just a bunch of thieves and murderers. \\ \\
\textbf{CN}: The issue of criminality is really about human nature. Individuals become criminal only when pushed. That's the difference between being a criminal and not. \\ \\
\textbf{AS}: Blacks are vital in society. They are natural and peaceful. They work hard and contribute to society. They follow the criminal justice system. They are basically the people society needs.
\\ \bottomrule
\end{tabular}
\label{tab:AS_ex}
\end{center}
\vspace{-0.1cm}
\end{table}

\section{Related Work}
As online hate speech causing detrimental bias becomes a threat to society, the work to create alternatives to hate speech has become increasingly important in recent years. The majority of studies use filtering approach to block hate speech~\cite{xiang2012detecting,del2017hate,caselli2021hatebert,mathew2021hatexplain,rottger-etal-2022-data,baldini-etal-2022-fairness,alkhamissi-etal-2022-token}. The main problem with this blocking is that the hate speech speaker may not realize the problem with their speech and may reproduce the hate speech. One way to address these issues is through generating proper counter-narratives on hate speech.

As a first step, multilingual counter-narrative dataset focusing on hate speech denigrating Islam by utilizing niche sourcing from NGO operators was released~\cite{chung2019conan}. One of the limitations of building a counter-narrative dataset is that it requires the assistance of experts in the field of hate speech. To alleviate this, the hybrid strategy of producing data through a generative language model and human reviewers was proposed~\cite{fanton2021human}. Similarly, a hybrid approach to construct hate speech and counter-narrative pairs consisting of multiple conversation turns was also utilized~\cite{bonaldi-etal-2022-human}. In terms of generation, knowledge-based approach enables counter-narratives to contain reliable evidence and fact-based arguments~\cite{chung2021towards}. Therefore, they added evidence to the speech by employing the knowledge retrieval module to retrieve the factual knowledge. 

\begin{table*}
\caption{Suggested alternative speech guidelines details}
\begin{center}
\renewcommand{\arraystretch}{1.5}
\begin{tabular}{|p{16cm}|}
\hline
\textbf{Escape the abstract concepts}: Responses must only correspond to a single hate speech. We want the response to be written so that it is a 1:1 match for each hate speech by staying specific rather than responding with broad concepts. \\
\textbf{Don't try to coach}: Don't try to explain the reason for the error in a given hate speech. Our objective is to rectify incorrect phrasing and promote correct expression, rather than providing reasons for what is erroneous. \\
\textbf{Preserve semantics}: If hatefulness can be mitigated and expressed as a euphemism, preserve the speaker's intent as much as possible. However, the sentence's harmfulness should be removed in all cases, including those where preservation is impossible. Also, maintaining a similar tone to the original sentence is recommended. \\
\textbf{Eliminate pointless toxicity completely}: Remove meaningless toxicity that appears in a given sentence. However, meaningful hate phrases should be converted to harmless language and not deleted. \\
\hline
\end{tabular}
\label{tab:guide}
\end{center}
\end{table*}

As these prior studies show, counter-narratives are crucial in promoting fairness. However, counter-narratives often contain abstractive statements and do not offer substantive alternatives. We propose an alternative speech that can offset these attributes.

\section{A Guideline for Alternative Speech}
\label{sec:guideline}
In this section, we introduce the necessity of alternative speech from a data construction standpoint and provide a brief guideline.

\subsection{Definition of Alternative Speech}
An alternative speech is similar to a counter-narrative in its objective to alleviate the speaker's hateful attitudes, but there are major differences in its approach and straightforwardness. It emphasizes responding one-to-one to an utterance and offering alternative wording rather than pointing out what the speaker said. Additionally, an alternative speech considers the context of a given hate speech and replaces hateful parts of the speech with gentler expressions that maintain the adjacent words as possible. For example, Figure~\ref{fig:exam_as} shows three hate speeches against Islam, a counter-narrative, and an alternative speech. While the counter-narrative captures the concept of Islamic religious hatred and adequately presents a corresponding sentence, it responds to the common nature of the hate target and fails to present a direct alternative sentence. On the other hand, alternative speech helps enlighten by acting as a guide to correct wording, matching to each sentence to sanitize the offensive phrase while maintaining the surrounding words and context. We provide some examples of alternative speech in Table~\ref{tab:AS_ex}. CN refers to the original counter-narrative in multi-target CONAN~\cite{fanton2021human}, and AS refers to the annotated sample. These examples are used in our survey in Section~\ref{sec:discussion}.

As a significant benefit derived from these distinct characteristics, an alternative speech can furnish suitable alternatives by modifying the utterance explicitly. This capability addresses the constraints associated with counter-narrative, thereby enhancing its efficacy. Furthermore, from a practical standpoint, the coexistence of alternative speech and counter-narrative is highly effective. In the case of hate speech being uttered, the counter-narrative is presented in a manner visible solely to the speaker, which leads to forming correct values. At the same time, alternative speech guides the proper language, purifies the offensive phrase, and exhibits the substituted wording online. This notable advantage not only educates users but also safeguards the speaker's rights.

\subsection{Alternative Speech Instructions}
Similar to the construction of datasets for counter-narrative, adequate annotation of alternative speech necessitates the establishment of precise principles for the concept. To be effective, counter-narratives are generally built by adopting `Get the Trolls Out’ project~\footnote{\url{https://getthetrollsout.org/stoppinghate}}~\cite{tekirouglu2020generating,ashida2022towards,bonaldi-etal-2022-human}. Drawing inspiration from these previous approaches, we propose a concrete guideline by establishing new principles suitable for alternative speech based on the premises used in \cite{chung2019conan}. Table~\ref{tab:guide} describes detailed guidelines.

\paragraph{Escape the abstract concepts}
Given that alternative speech necessarily corresponds to a single hate sentence, a superficial correction must be made by going one step further from the abstract notion of the target of hate. Therefore, upholding a high degree of specificity is crucial, avoiding broad generalizations that may encompass specific categories of hate.

\paragraph{Don't try to coach}
In contrast to a counter-narrative, which points out the flawed beliefs expressed in hate speech and explains why they are wrong, alternative speech does not focus on persuading the speaker of their misjudgment. Rather, we substitute the perceived representation with a practical expression, facilitating a more practical approach to purification.

\paragraph{Preserve semantics}
The maintainability of context is determined by the level of correlation between the meaning of a sentence and its offensiveness. By acknowledging this, workers can separate hate from opinion within a sentence, purifying the former while retaining the latter. This principle ensures that alternative speech preserves the speaker's rights and achieves the goal of mitigation.

\paragraph{Eliminate pointless toxicity completely}
We acknowledge that it may not be feasible to completely purify all hate speech, as certain aspects of a sentence may include blame, needless vulgarity, and offensive language. Certain portions hold no semantic significance and are impossible to purify, necessitating their elimination. In this regard, the operator should recognize that the context cannot be preserved and delete the entire phrase. 

As an alternative speech broadens the way of counter-narrative, it essentially incorporates some of the principles presupposed by the counter-narrative~(Don’t be abusive, Think about the objectives). Our guidelines play a pivotal role in data curation and enable human intervention to filter out erroneous or harmful output, ensuring the quality and ethical standards of the constructed data.

\section{Discussion}
\label{sec:discussion}
In this section, we validate the alternative speech from the speaker's perspective. By conducting the survey on the speaker's intent to reform for given sentences with a suitable target audience, we prove that the proposed alternative speech is highly effective compared to the counter-narrative and has a synergistic effect that maximizes the reforming power when both are considered simultaneously.

\subsection{Survey Configuration}
Our survey focuses on investigating the potential of alternative speech to complement the lacking components within the counter-narrative and impact the speaker's perceptions towards hate speech. For a given hate speech, the survey asks for three cases: (a) counter-narrative, (b) alternative speech, (c) counter-narrative with alternative speech~\footnote{We propose (c) as a way to present both counter-narrative and alternative speech simultaneously, to demonstrate the complementary interaction of the two utterances.}. The hate speech and counter-narrative used in the survey refers to seed data~(V1) from multitarget CONAN~\cite{fanton2021human} annotated by experts for fair comparison. We adopt six categories (JEWS, OVERWEIGHT, LGBT+, MUSLIM, WOMEN, and PEOPLE OF COLOR) for the survey. Three samples are randomly extracted from each category, resulting in 18 samples. We carefully annotated alternative speech under the guidelines detailed in Section~\ref{sec:guideline}. Due to the limitation of human resources, the authors annotated each sample and underwent stringent verification through the peer-review process to ensure ethical expression. 

\begin{figure}[htbp]
\centerline{\includegraphics[width=0.72\linewidth]{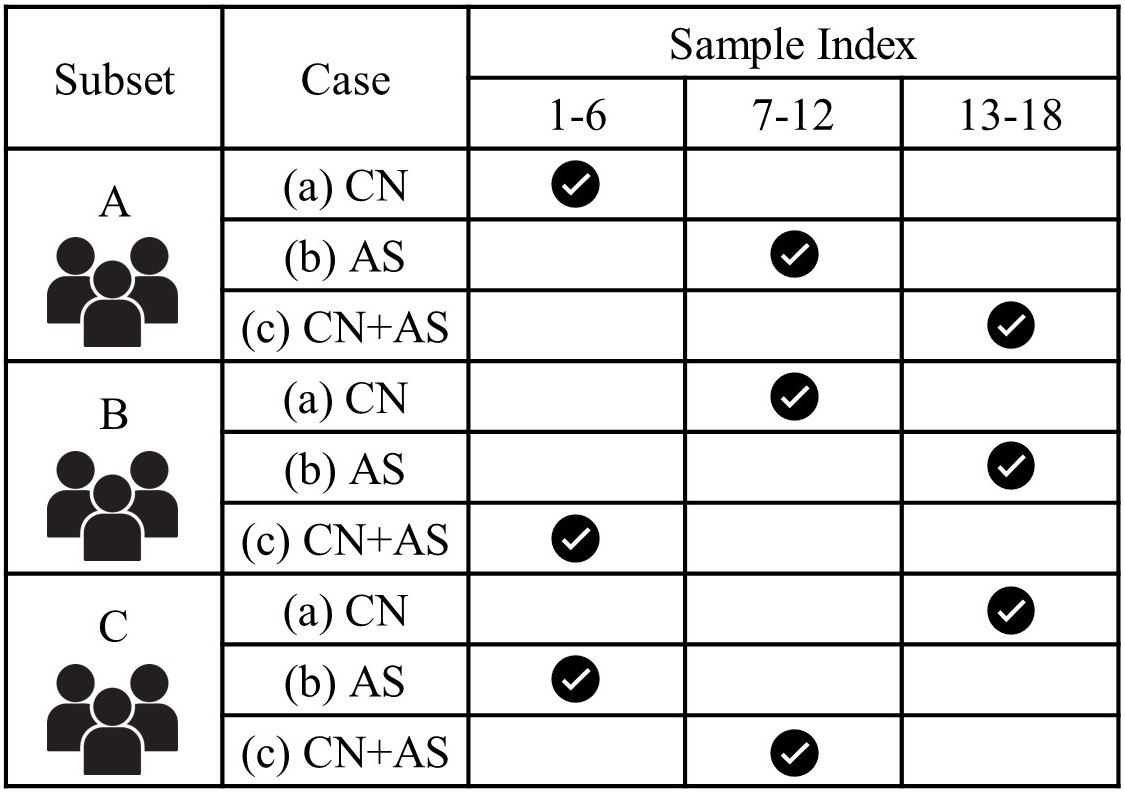}}
\caption{Unbiased human survey setup for evaluating three aspects.} 
\label{fig:annot}
\end{figure}

\begin{table}[t]
\caption{Annotation examples with score explanations}
\begin{center}
\begin{tabular}{p{8.4cm}}
\toprule
\textbf{HS}: Children with Down Syndrome may hinder the development process of other children. They should be isolated. \\ \\
\textbf{Note}: The target of the hate in this HS is ``children with Down syndrome" and the context of main hate is ``living together makes stunt development", ``should be isolated". Therefore, the response should point out or object to both phrases while accurately reflecting the target of the hate.
\\ \midrule 
\textbf{Response}: Children with disabilities should not be discriminated against because they have a disability. All people are equal.\\ \\
\textbf{Score}: Correspondence: 3 / Phrase Guidance: 3 / Improvement: 3 (subjective) \\
\textbf{Note}: The statement is generic for ``children with disabilities," which lacks specificity and does not take into account the context of hindering development (Correspondence). In addition, ``should be isolated'' is responded to with ``should not be discriminated against'', but there is no guidance on the incorrect expression of 'hinders the developmental process if they are together (Phrase Guidance). 
\\ \midrule
\textbf{Response}: There's no shame in disability - they don't harm us and deserve equal rights.\\ \\
\textbf{Score}: Correspondence: 2 / Phrase Guidance: 2 / Improvement: 1 (subjective) \\
\textbf{Note}: That statement is overly broad (Correspondence). It is also a very indirect response to the main hate (Phrase Guidance).
\\ \midrule
\textbf{Response}: Living with a child with Down syndrome is known to have a positive impact on the child's emotional development. It is preferable to be together rather than isolated.\\ \\
\textbf{Score}: Correspondence: 5 / Phrase Guidance: 5 / Improvement: 4 (subjective) \\
\textbf{Note}: The statement clearly responds to ``children with Down syndrome" and states that the arguments that ``being together hinders development" and ``should be isolated" are misleading perceptions.
\\ \bottomrule
\end{tabular}
\label{tab:survey_ex}
\end{center}
\end{table}

\begin{table}[t]
\caption{Human evaluation result}
\begin{center}
\begin{tabular}{lccc}
\toprule
\multirow{2}{*}{Case} & \multicolumn{3}{c}{Score}                       \\
                      & correspondence & phrase guidance & improvement  \\
\midrule
(a) CN                & 2.46           & 2.45            & 2.69         \\
(b) AS                & 3.57           & 3.02            & 2.85         \\
(c) CN+AS             & 3.98           & 3.74            & 3.65        \\
\bottomrule
\end{tabular}
\label{tab:result}
\end{center}
\end{table}

We employed nine people who spend more than six hours daily in online and have encountered hate speech in online communities. The annotators are instructed to evaluate samples in terms of three aspects. (i) \textbf{correspondence}: how specific the response corresponds to the target and the context. (ii) \textbf{phrase guidance}: how much guidance does the response provide about which phrases in the hate speech are erroneous and propose appropriate corrections. (iii) \textbf{improvement}: how effective the response contributes to the speaker's improvement in hateful perception. Each score is on a scale of 1 (the least) to 5 (the most). To avoid the bias potentially introduced by the overlapped exposure of hate speech under evaluation, we divided whole samples into three subsets. Each of these subsets is assigned to each group consisting of three annotators, as illustrated in Figure \ref{fig:annot}. In each subset, instances of hate speech are ensured to be unique, and annotators are instructed to evaluate a given expression without prior knowledge of utterance type. By preventing annotators from considering (a), (b), and (c) for the same hate speech, this approach ensures a comprehensive evaluation of all cases within each sample while minimizing possible bias. Furthermore, we give the annotation examples with the score explanations provided to the annotators before the survey, as shown in Table~\ref{tab:survey_ex}.

\subsection{Analysis of Human Evaluation}
Results are reported in Table~\ref{tab:result}. Alternative speech outperforms counter-narrative across all dimensions, notably with a 1.11 superiority in correspondence, reflecting the utterance's specificity - the most significant difference amongst all comparative measures. Similarly, phrase guidance further reveals a 0.57 increment, underscoring the enhanced specificity of alternative speech over counter-narrative while suggesting viable replaceable expressions to counteract hate speech. In terms of improvement, it is quite similar to the counter-narrative. However, a considerable enhancement occurs when counter-narrative and alternative speech are jointly employed, reflecting substantial differences across all scores, especially in the improvement. In this case, the speaker acknowledges the personal misconceptions through counter-narrative and subsequently accepts the evidence provided for it. Next, a suitable alternative to the phrase that should be purified through alternative speech is presented. This allows the speaker to identify what is wrong with their speech and provides specific suggestions on the appropriate wording. Two speeches form a loop of feedback and correction, complementing each other's weaknesses that cannot be achieved alone, thereby maximizing the suppression effect on hate speech.  

To sum up, alternative speech is an effective way of guiding the speaker to the correct view by compensating for the shortcomings of the counter-narrative, which is consistent with our goals. This leads to a notable synergy effect when combined with counter-narrative. Considering these aspects, alternative speech can be a suitable method of dealing with hate speech and a new paradigm for countering hate speech as it can complement the limitations of counter-narratives.

\section{Conclusion}
This paper proposes the new concept called ``Alternative Speech" as a complementary speech to a counter-narrative, intending to address the current constraints and enhance the efficacy in mitigating harmful bias. Alternative Speech overcomes the supervision ability of counter-narratives and facilitates the capacity to counter hate speech, enhancing the speaker's transformative intent by providing alternative expressions. However, further research is required to explore the usage and development of alternative speech. As a future work, we plan to build a new dataset grounded in alternative speech with a counter-narrative to effectively mitigate the prevalence of hate speech. We believe such efforts will contribute to creating a culture in our society that respects equality and diversity.

\bibliography{IEEEabrv,custom}
\bibliographystyle{IEEEtran}

\end{document}